\documentclass[draft]{agujournal2019}

\usepackage{url} %this package should fix any errors with URLs in refs.
\usepackage[inline]{trackchanges} %for better track changes. finalnew option will compile document with changes incorporated.
\usepackage{soul}
\usepackage{booktabs}
\usepackage{todonotes}
\usepackage{cleveref}
\usepackage{subfig}
\draftfalse

\journalname{arXiV}

\begin{document}

%%%%%%%%%%%%%%%%%%%%%%%%%%%%%%%%%%%%%%%%%%%%%%%
%  TITLE
%
% (A title should be specific, informative, and brief. Use
% abbreviations only if they are defined in the abstract. Titles that
% start with general keywords then specific terms are optimized in
% searches)
%
%%%%%%%%%%%%%%%%%%%%%%%%%%%%%%%%%%%%%%%%%%%%%%%

% Example: \title{This is a test title}
\title{ERA5-Derived Data and Resources for Generative Modeling of 2-m Air Temperature over South Asia}
%%%%%%%%%%%%%%%%%%%%%%%%%%%%%%%%%%%%%%%%%%%%%%%
%
%  AUTHORS AND AFFILIATIONS
%
%%%%%%%%%%%%%%%%%%%%%%%%%%%%%%%%%%%%%%%%%%%%%%%

% Authors are individuals who have significantly contributed to the
% research and preparation of the article. Group authors are allowed, if
% each author in the group is separately identified in an appendix.)

% List authors by first name or initial followed by last name and
% separated by commas. Use \affil{} to number affiliations, and
% \thanks{} for author notes.
% Additional author notes should be indicated with \thanks{} (for
% example, for current addresses).

% Example: \authors{A. B. Author\affil{1}\thanks{Current address, Antartica}, B. C. Author\affil{2,3}, and D. E.
% Author\affil{3,4}\thanks{Also funded by Monsanto.}}

\authors{Siddharth Rout\affil{1,2}}
% \affil{1}, St\'ephane Gaudreault\affil{2}, Shoyon Panday\affil{2}}

\affiliation{1}{University of British Columbia, 6200 University Blvd, Vancouver, BC V6T 1Z4, Canada}
\affiliation{2}{Northeastern University, 410 W Georgia St \#1400, Vancouver, BC V6B 1Z3, Canada}
% \affiliation{2}{Environment and Climate Change Canada, 2121 Trans-Canada Hwy, Dorval, Qu\'ebec H9P 1J3, Canada}
% \affiliation{4}{Fourth Affiliation}

%(repeat as many times as is necessary)

% Corresponding author mailing address and e-mail address:

% (include name and email addresses of the corresponding author.  More
% than one corresponding author is allowed in this LaTeX file and for
% publication; but only one corresponding author is allowed in our
% editorial system.)

% Example: \correspondingauthor{First and Last Name}{email@address.edu}

% \correspondingauthor{Siddharth Rout}{s.rout@northeastern.edu}

%%%%%%%%%%%%%%%%%%%%%%%%%%%%%%%%%%%%%%%%%%%%%%%
% KEY POINTS
%%%%%%%%%%%%%%%%%%%%%%%%%%%%%%%%%%%%%%%%%%%%%%%
%  List up to three key points (at least one is required)
%  Key Points summarize the main points and conclusions of the article
%  Each must be 140 characters or fewer with no special characters or punctuation and must be complete sentences

% Example:
% \begin{keypoints}
% \item	List up to three key points (at least one is required)
% \item	Key Points summarize the main points and conclusions of the article
% \item	Each must be 140 characters or fewer with no special characters or punctuation and must be complete sentences
% \end{keypoints}

\begin{keypoints}
\item A processed regional subset of ERA5 and associated ML resources for 40-year (1985--2024), 3-hourly, 0.25$^{\circ}$ gridded 2-m air temperature field provides a consistent regional record across South Asia and surrounding regions.
\item The dataset is prepared for reproducible machine-learning research with standardized data, temporal conditioning, predefined training and testing periods, and precomputed climatological statistics.
\item Generative flow matching demonstrates the dataset's utility for generating spatially coherent temperature perturbations and temperature fields conditioned on day of year and time of day.
\end{keypoints}


%%%%%%%%%%%%%%%%%%%%%%%%%%%%%%%%%%%%%%%%%%%%%%%
%
%  ABSTRACT and PLAIN LANGUAGE SUMMARY
%
% A good Abstract will begin with a short description of the problem
% being addressed, briefly describe the new data or analyses, then
% briefly states the main conclusion(s) and how they are supported and
% uncertainties.

% The Plain Language Summary should be written for a broad audience,
% including journalists and the science-interested public, that will not have 
% a background in your field.
%
% A Plain Language Summary is required in GRL, JGR: Planets, JGR: Biogeosciences,
% JGR: Oceans, G-Cubed, Reviews of Geophysics, and JAMES.
% see http://sharingscience.agu.org/creating-plain-language-summary/)
%
%%%%%%%%%%%%%%%%%%%%%%%%%%%%%%%%%%%%%%%%%%%%%%%

%% \begin{abstract} starts the second page

\begin{abstract}
Reliable, high-resolution temperature datasets are essential for characterizing regional climate variability, developing data-driven weather and climate applications, and benchmarking machine-learning methods for geophysical data using personal computers or workstations. We present machine learning resources and applications for a 40-year, 3-hourly, 0.25° gridded sub-sampled dataset of 2-m air temperature derived from ERA5 over South Asia and surrounding regions for 1985--2024. The dataset provides spatially and temporally consistent near-surface temperature fields. With machine-learning-ready preprocessing, including temporal conditioning by day of year and 3-hourly time of day, standardized data representation, and predefined training and testing periods, these characteristics facilitate reproducible experimentation with spatiotemporal learning and generative modeling while preserving seasonal and diurnal temperature variability. To demonstrate the utility of the dataset for artificial intelligence applications, we employ deterministic generative flow matching to generate spatially coherent temperature perturbations and temperature fields conditioned on day of year and time of day. The resulting fields are evaluated against the corresponding ERA5 temperature statistics and spatial characteristics. To facilitate reproducibility and reuse, we provide the processed dataset, training and generation code, and precomputed climatological mean and standard deviation fields for each day of the year and 3-hourly time of day. Together, these resources provide an accessible ML-ready framework for research in spatiotemporal machine learning, generative modeling, climate-data analysis, uncertainty quantification, and regional temperature variability across South Asia and surrounding regions. Data and code are available at: \url{https://www.kaggle.com/datasets/siddharthrout/indian-subcontinent-temperature-field}.
\end{abstract}

\section*{Plain Language Summary}
ERA5 is a large global reanalysis dataset that provides detailed information about the Earth's atmosphere, but its size and complexity can make it difficult to use directly for research and experimentation on personal computers and in small university computing environments. We provide a more accessible, machine-learning-ready dataset of 2-m air temperature derived from ERA5 over the South Asian subcontinent and surrounding regions for 1985--2024. The dataset has a spatial resolution of 0.25$^{\circ}$ and includes temperature fields at 3-hour intervals, allowing temperature variations to be examined across individual grid cells, seasons, days of the year, and times of day. The region includes diverse and climatically important areas such as the Tibetan Plateau, Himalayan Range, Persian Gulf, Southeast Asia, Arabian Sea, Bay of Bengal, and northern Indian Ocean. We also demonstrate how the dataset can be used to generate additional temperature information, including spatially coherent temperature perturbations and temperature fields conditioned on any day of the year and 3-hourly time of day. By providing a compact, consistently processed dataset together with analysis and generative-modeling resources, this work lowers the computational and preprocessing barriers to experimentation with weather and climate data. The dataset can support research on regional climate variability and extremes, climate-model evaluation, uncertainty quantification, and the development and benchmarking of artificial-intelligence methods for weather and climate applications. The resulting dataset and associated resources provide a reusable foundation for studying temperature variability across the diverse environments and densely populated regions of South Asia and surrounding areas.

%%%%%%%%%%%%%%%%%%%%%%%%%%%%%%%%%%%%%%%%%%%%%%%
%
%  BODY TEXT
%
%%%%%%%%%%%%%%%%%%%%%%%%%%%%%%%%%%%%%%%%%%%%%%%

%%% Suggested section heads:
% \section{Introduction}
%
% The main text should start with an introduction. Except for short
% manuscripts (such as comments and replies), the text should be divided
% into sections, each with its own heading.

% Headings should be sentence fragments and do not begin with a
% lowercase letter or number. Examples of good headings are:

% \section{Materials and Methods}
% Here is text on Materials and Methods.
%
% \subsection{A descriptive heading about methods}
% More about Methods.
%
% \section{Data} (Or section title might be a descriptive heading about data)
%
% \section{Results} (Or section title might be a descriptive heading about the
% results)
%
% \section{Conclusions}

\section{Introduction}

Near-surface air temperature is a fundamental variable for characterizing climate variability and extremes, evaluating weather and climate models, and developing data-driven applications \cite{lindzen1987role,qin2001mono}. South Asia exhibits pronounced spatial and temporal temperature variability due to its broad latitudinal extent, complex topography, heterogeneous land-surface characteristics, monsoon circulation, and interactions with the surrounding oceans \cite{arora2005evaluation,roxy2015drying}. Long-term, spatially consistent temperature datasets are therefore valuable for regional climate analysis and for applications requiring realistic representations of temperature fields. Observed warming and changes in temperature extremes across Asia have further increased the need for reliable regional temperature information for climate-impact studies \cite{portner2022climate}.

Reanalysis products provide spatially and temporally consistent estimates of historical atmospheric conditions by combining observations with numerical weather prediction models through data assimilation \cite{dee2011era,law2015data}. The fifth-generation European Centre for Medium-Range Weather Forecasts (ECMWF) reanalysis, ERA5, provides hourly global estimates of atmospheric and surface variables from 1940 onward \cite{hersbach2020era5}. Its 2-m air temperature product is widely used in weather and climate studies and has also been adopted in recent data-driven weather and climate applications \cite{price2025probabilistic,bonev2025fourcastnet}. However, extracting, subsetting, and preprocessing large global reanalysis archives for regional studies and machine-learning applications can be computationally and operationally demanding. A dedicated regional dataset can reduce these preprocessing requirements, facilitate reproducible machine-learning workflows, and provide an accessible testbed for the development and evaluation of artificial intelligence for science (AI4Sc).

The development of data-driven methods for weather and climate applications also requires datasets that preserve the spatial and temporal structure of atmospheric fields. Independent random perturbations, for example, generally do not reproduce the spatial correlations and coherent structures characteristic of atmospheric temperature fields, limiting their usefulness for applications such as ensemble generation and uncertainty quantification \cite{toth1993ensemble}. Generative artificial intelligence (GenAI) methods provide an alternative by learning statistical relationships from data and generating new samples that reflect the underlying data distribution \cite{feuerriegel2024generative}. Recent diffusion- and flow-based approaches have demonstrated potential for weather and climate applications, including probabilistic prediction, ensemble generation, uncertainty quantification, and downscaling \cite{price2025probabilistic,mardani2025residual}. More generally, generative models can learn distributions of atmospheric fields and produce spatially structured realizations that retain important characteristics of the training data \cite{rout2025fast,rout7607514flow}. Flow matching \cite{lipman2022flow} provides a framework for learning continuous transformations between a reference distribution and a target data distribution. Recent studies have investigated flow matching for atmospheric and weather-related generation tasks, including applications at major operational forecasting centres such as the European Centre for Medium-Range Weather Forecasts (ECMWF) \cite{Lang2026}.

For temperature data, seasonal and diurnal variability represent important sources of temporal structure \cite{vinnikov2002diurnal,vinnikov2004analysis}. Characterizing temperature statistics as a function of day of year and time of day provides a compact representation of these dependencies and can support both climatological analysis and conditional generative modeling. Such conditioning also provides a practical way to generate temperature fields that are consistent with the expected seasonal and diurnal state of the atmosphere.

In this study, we sub-sample a 40-year, 3-hourly, 0.25$^{\circ}$ gridded dataset of 2-m air temperature derived from ERA5 over South Asia and surrounding regions for 1985--2024. The dataset is organized and processed to facilitate reproducible machine-learning research, including temporal conditioning by day of year and 3-hourly time of day, predefined training and testing periods, and precomputed climatological statistics. We further demonstrate the utility of the dataset through a generative flow-matching framework that generates spatially coherent temperature perturbations and temperature fields conditioned on day of year and time of day. The dataset and associated resources are intended to provide an accessible foundation for research in spatiotemporal machine learning, generative modeling, climate-data analysis, and regional temperature variability.

\section{Data and Dataset Description}

We used a 40-year, 3-hourly gridded dataset of 2-m air temperature over South Asia and surrounding regions for 1 January 1985--31 December 2024. The dataset is derived from the ECMWF ERA5 reanalysis \cite{hersbach2020era5} and is designed to provide a compact, consistently processed resource for regional climate analysis and reproducible machine-learning research. The dataset contains temperature fields in Kelvin ($K$) on a $257\times257$ grid with a spatial resolution of $0.25^\circ\times0.25^\circ$, covering $10^\circ$S--$54^\circ$N and $50^\circ$E--$114^\circ$E. It contains 116,880 temporal fields, with 66,049 grid cells per field, corresponding to approximately 7.72 billion individual temperature values.

The temperature fields are provided as PyTorch \texttt{.pt} files to facilitate direct integration with machine-learning workflows. In addition to the historical temperature fields, the dataset includes precomputed climatological mean and standard deviation fields conditioned on day of year and 3-hourly time of day. These statistics provide an efficient representation of seasonal and diurnal temperature variability and can be used for analysis, normalization, and conditional generative modeling without repeatedly processing the full historical archive. The dataset is therefore intended not only as a regional temperature archive, but also as an ML-ready resource for reproducible experimentation with spatiotemporal and generative models.

For the generative-model experiments presented in this study, data from 1985--2019 are used for training, while data from 2020--2024 are reserved for testing. This chronological separation provides an out-of-sample evaluation period that is temporally distinct from the training data. Prior to model input, temperature is nondimensionalized by subtracting an approximate reference mean of 290 $K$ and dividing by an approximate reference standard deviation of 10 $K$. The same transformation is applied consistently to the training and testing data. Detailed dataset characteristics are summarized in \Cref{tab:datadesc}.

\begin{table}[t]
\centering
\caption{Specifications of the regional 2-m air temperature dataset.}
\label{tab:datadesc}
\begin{tabular}{ll}
\hline
\textbf{Property} & \textbf{Description} \\
\hline
Variable & 2-m air temperature \\
Data source & ECMWF ERA5 reanalysis \\
Period & 1 January 1985--31 December 2024 \\
Temporal resolution & 3 hourly \\
Spatial grid & $257 \times 257$ \\
Spatial resolution & $0.25^{\circ} \times 0.25^{\circ}$ \\
Latitude domain & $10^{\circ}$S--$54^{\circ}$N \\
Longitude domain & $50^{\circ}$E--$114^{\circ}$E \\
Units & Kelvin ($K$) \\
Number of temporal fields & 116,880 \\
Grid cells per field & 66,049 \\
Total grid-cell values & 7,719,807,120 \\
Storage format & PyTorch \texttt{.pt} \\
Datatype & PyTorch float32 \\
Storage volume & 30.97 GB \\
\hline
\end{tabular}
\end{table}

\subsection{Region Covered}

The geographic domain extends from $10^\circ$S to $54^\circ$N and $50^\circ$E to $114^\circ$E, encompassing a diverse range of climatic and geographic environments across South Asia and surrounding regions. The domain includes parts of the Middle East and Arabian Peninsula, Central and South Asia, the Himalayan--Tibetan region, Southeast Asia, and western and central China, as well as the Arabian Sea, Bay of Bengal, and northern Indian Ocean.

The region contains substantial contrasts in climate, elevation, land-surface characteristics, and population density. These include the arid and semi-arid environments of the Middle East and Central Asia, the high-elevation Himalayas and Tibetan Plateau, the densely populated Indo-Gangetic Plain, tropical and subtropical environments of Southeast Asia, and the major oceanic regions that influence the Asian monsoon system. These contrasts provide a diverse setting for examining spatial and temporal temperature variability within a common $0.25^\circ$ gridded framework.

Representative population centers within the geographic domain include Kabul (Afghanistan), Dhaka (Bangladesh), Thimphu (Bhutan), Phnom Penh (Cambodia), Urumqi and Chongqing (China), New Delhi, Mumbai, Kolkata, Bengaluru, Chennai, and Hyderabad (India), Jakarta (Indonesia), Tehran (Iran), Astana (Kazakhstan), Bishkek (Kyrgyzstan), Vientiane (Laos), Kuala Lumpur (Malaysia), Yangon (Myanmar), Kathmandu (Nepal), Muscat (Oman), Islamabad and Lahore (Pakistan), Doha (Qatar), Colombo (Sri Lanka), Singapore (Singapore), Bangkok (Thailand), Dushanbe (Tajikistan), Ashgabat (Turkmenistan), Dubai (United Arab Emirates), Tashkent (Uzbekistan), and Hanoi (Vietnam). These locations illustrate the broad geographic and climatic diversity represented within the dataset.

\section{Dataset Applications}

The dataset provides a reusable resource for studying regional temperature variability, seasonal and diurnal temperature patterns, temperature extremes, and spatial temperature structures. Its consistent spatial and temporal organization facilitates reproducible analysis across a wide range of machine-learning and climate-data applications. The precomputed climatological statistics further enable efficient analysis of temperature variability without requiring repeated processing of the complete 40-year archive.

The dataset also provides a testbed for generative modeling of spatially structured temperature fields. As a demonstration, we use flow matching to generate spatially coherent temperature perturbations and temperature fields conditioned on day of year and 3-hourly time of day. Such generated fields can support applications including data augmentation, ensemble generation, sensitivity experiments, and uncertainty-quantification studies. These applications should be understood as potential uses of the dataset and demonstrated capabilities of the generative framework rather than as forecasts of future atmospheric states.

More broadly, the dataset can support the development and benchmarking of artificial-intelligence methods for weather and climate science, including spatiotemporal representation learning, conditional generative modeling, temperature-field generation, and regional climate analysis. By providing consistently processed data together with machine-learning-oriented metadata, climatological statistics, and associated analysis and modeling resources, the dataset reduces the preprocessing burden associated with working directly with large global reanalysis archives.

\section{Generative Modeling Demonstration}

GenAI models are increasingly being applied in atmospheric and climate sciences to represent complex data distributions, generate synthetic realizations, and characterize variability and uncertainty. Approaches including diffusion models \cite{nichol2021improved}, generative adversarial networks (GANs) \cite{goodfellow2014generative}, and neural operators \cite{kovachki2023neural,bonev2025fourcastnet} have been used to learn patterns from observations, reanalysis products, and numerical weather simulations. These approaches have been applied to tasks such as high-resolution weather generation, statistical downscaling, data augmentation, and probabilistic prediction. These models can be viewed as learning the underlying probability distribution represented by a dataset and subsequently using the learned distribution to generate new samples. For high-dimensional atmospheric fields, this distribution contains both marginal variability at individual grid points and spatial dependencies between locations. Preserving these dependencies is particularly important when generating temperature fields, since atmospheric temperature variability exhibits coherent spatial structures that cannot generally be reproduced by independently perturbing individual grid cells.

\begin{figure}[t]
    \centering
    \includegraphics[width=0.48\textwidth]{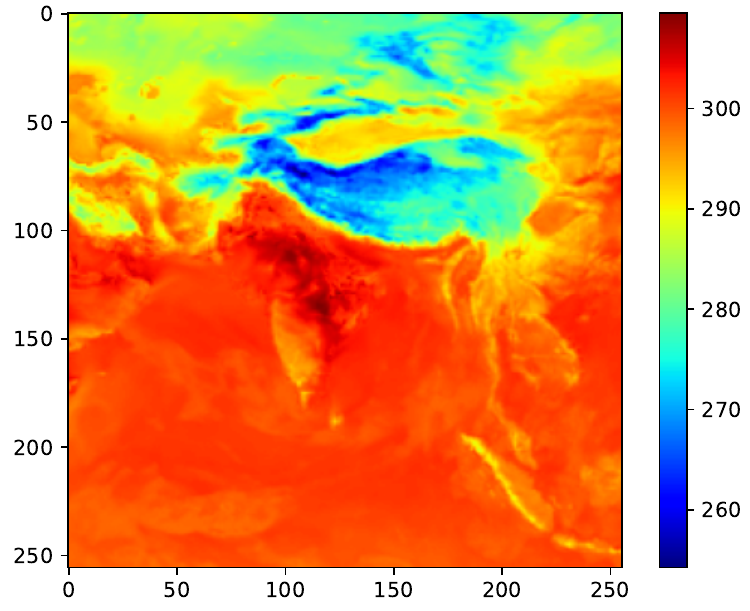}
    \hfill
    \includegraphics[width=0.48\textwidth]{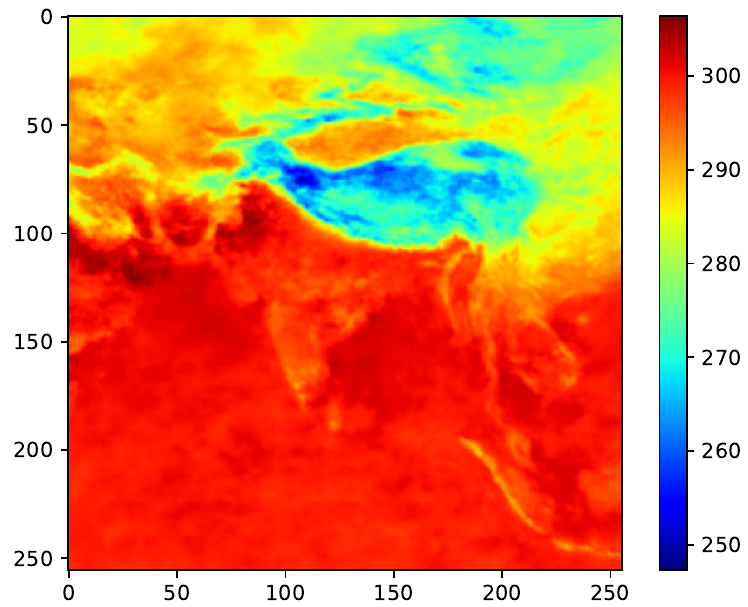}
    \includegraphics[width=0.8\textwidth]{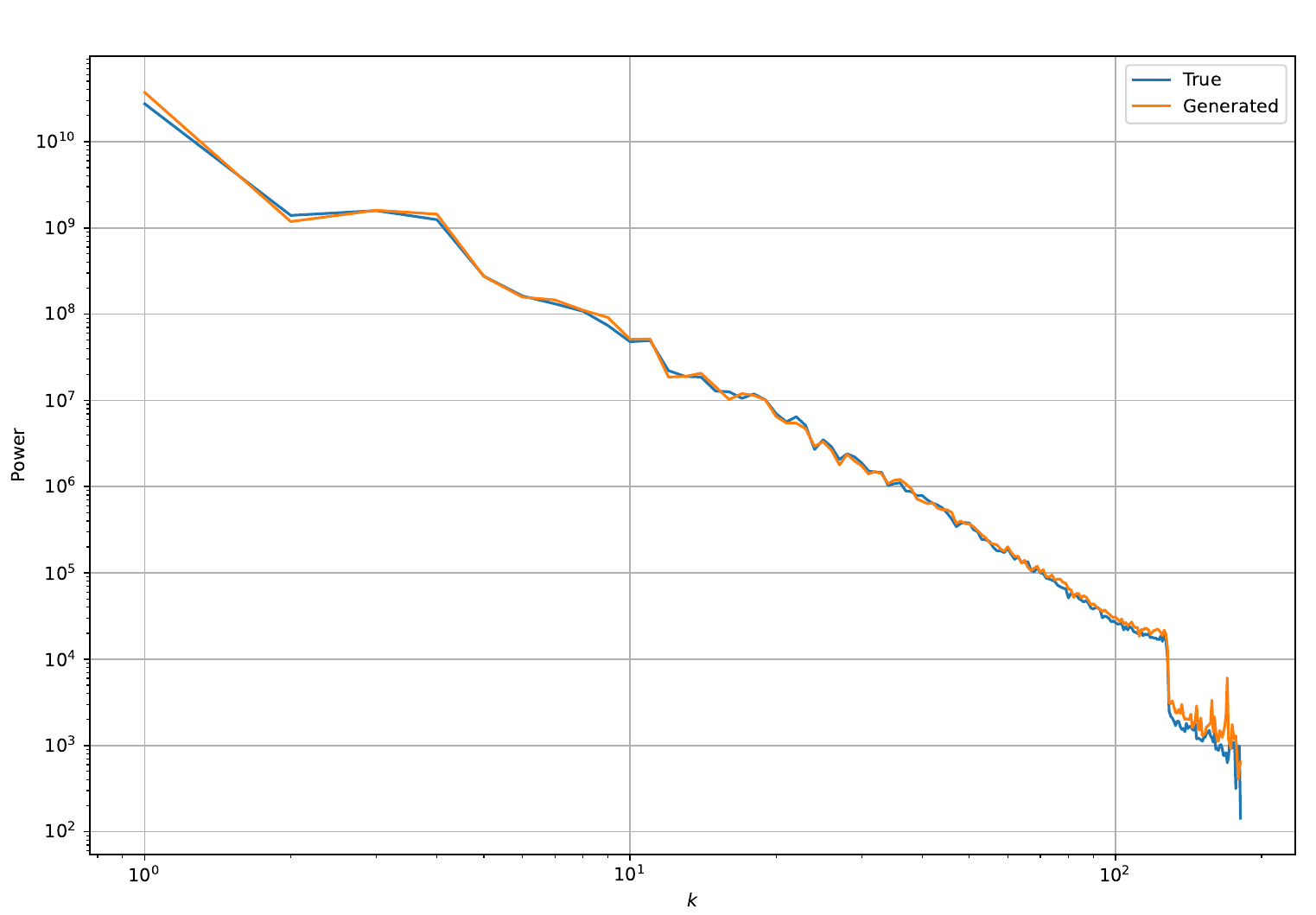}
    \caption{A true reanalysis plot of temperature field in $K$ (top left) and a randomly generated temperature field in $K$ for the same day of the year and same hour of the day (top right). The power spectral density curves of temperature vs wavenumber($k$) for both these cases are very similar (bottom).}
    \label{fig:generated_sample}
\end{figure}

To evaluate the ability of a generative model to reproduce the spatial characteristics of temperature variability, we trained a flow matching model on temperature fields and subsequently sampled new realizations from the learned distribution. The generated fields were then compared with ERA5 reanalysis temperature fields in both physical and spectral space, for reference. \Cref{fig:generated_sample} shows an example comparison, with the existing reanalysis temperature field and a randomly generated temperature realization for the same day-of-the-year (154) and same time of the day (00UTC) as of the reference sample shown alongside their corresponding power spectral density \cite[PSD]{welch1967use} as a function of wavenumber. The PSD provides a quantitative comparison of the distribution of variability across spatial scales. The similarity between the spectral characteristics of the reference and generated fields indicates that the model is able to reproduce important scale-dependent features of the temperature perturbations, providing evidence that the learned generative process captures aspects of the spatial structure present in the historical data.

Flow matching provides a generative framework for learning a continuous transformation between a simple reference distribution and a target data distribution \cite{lipman2022flow}. In this study, we use a flow-matching model to learn the spatial structure of temperature perturbations derived from the historical temperature dataset. The objective is to generate additional realizations that retain the spatial organization of temperature variability represented in the ERA5. The framework, therefore, provides a data-driven approach for investigating and reproducing aspects of temperature variability that would be difficult to capture using independent random perturbations.

\subsection{Diffusion Transformer Architecture}

We employ a Diffusion Transformer (DiT) architecture \cite{peebles2023scalable} as the neural backbone for the flow-matching framework. The model operates directly on the gridded temperature fields and learns the velocity field required to transform samples from a simple reference distribution into the distribution of historical temperature fields. Unlike convolutional architectures, the Transformer-based formulation uses self-attention \cite{vaswani2017attention,dosovitskiy2020image} to model interactions between spatial locations, allowing the network to capture long-range spatial dependencies in temperature fields \cite{zakariaei2024advection,pereira2026learning}.

Images passed to vision transformers are often resized to resolutions that are powers of two, making them easier to partition into patches efficiently. So, resized $256\times256$ temperature fields are represented as spatial tokens before being processed by the DiT. The DiT is reconfigured to allow two conditional inputs. The details of the configurations can be seen in \Cref{tab:dit2c_config}. A continuous flow-time variable is embedded and provided to the network to condition the predicted velocity field on the position along the generative trajectory. For conditional generation, day-of-year and 3-hourly time-of-day information are additionally encoded and incorporated into the DiT. The model is trained using the flow-matching objective, with historical temperature fields providing samples from the target distribution and samples from a reference distribution providing the initial state of the generative process. At inference time, the learned velocity field is integrated to generate new temperature realizations. The same architecture is used for both spatially coherent perturbation generation and conditionally sampled temperature variability. The complete code can be found in the repository.

\begin{table}[t]
\centering
\caption{DiT model architecture and hyperparameter configuration.}
\label{tab:dit2c_config}
\begin{tabular}{lc}
\hline
\textbf{Parameter} & \textbf{Value} \\
\hline
Input image size & $256 \times 256$ \\
Number of input channels & 1 \\
Number of classes for first condition & 366 \\
Number of classes for second condition & 24 \\
Patch size & $16 \times 16$ \\
Transformer depth & 6 \\
Hidden dimension & 384 \\
Number of attention heads & 8 \\
Output image size & $256 \times 256$ \\
Number of output channels & 1 \\
\hline
\end{tabular}
\end{table}
\subsection{Realistic Temperature Perturbations for Weather Variability}

The historical temperature fields provide an empirical basis for learning the spatial structure of temperature variability and generating additional realizations beyond those contained in the original record. In this study, following \cite{rout2025fast}, flow matching is used to learn the distribution of temperature fields and generate spatially coherent perturbed fields. Unlike independently sampled random noise, which generally does not preserve the spatial dependence of atmospheric temperature variability, the flow-matching model learns spatial relationships directly from the historical temperature fields. See \Cref{fig:generated_perturbed_sample}, we take the same 154th day-of-the-year at 00 UTC. We perturb the whole field using the perturbation parameter $\sigma$, mentioned by Rout et al., of various amplitude. Larger is the $\sigma$, larger is the perturbation.

The resulting perturbed fields can be used as additional temperature realizations. Because the perturbations are generated as spatially distributed fields rather than as independent values at individual grid points, they retain spatial dependencies learned from the historical dataset. This provides a mechanism for increasing the number and diversity of available temperature realizations while preserving spatial organization characteristic of the observed temperature variability.

Such realizations can support applications requiring multiple spatially structured temperature fields, including ensemble generation, uncertainty characterization, sensitivity experiments, and machine-learning data augmentation. Importantly, the generated fields represent statistical realizations of temperature variability learned from the historical dataset rather than forecasts of specific future atmospheric states. Their usefulness, therefore, depends on the extent to which the learned distribution reproduces the relevant statistical and spatial characteristics of the underlying temperature data.

\clearpage

\begin{figure}[t]
    \centering
    \includegraphics[width=0.48\textwidth]{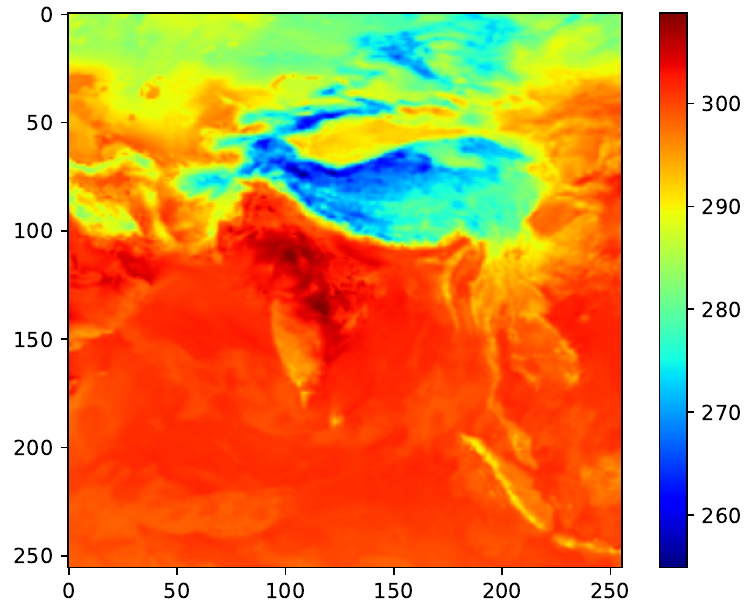}
    \hfill
    \includegraphics[width=0.48\textwidth]{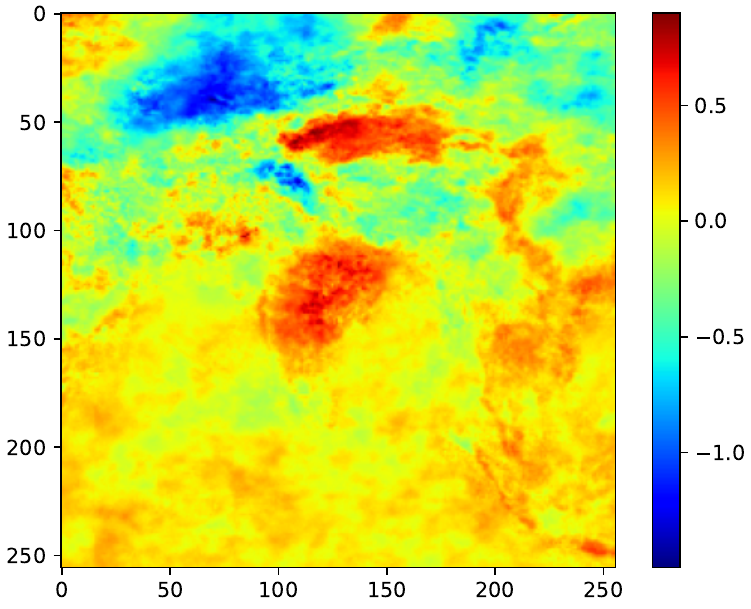}
    \caption{A perturbed reanalysis plot of temperature field in $K$ (left) and temperature perturbation field in $K$ for the 154th day-of-the-year at 00 UTC (right) for perturbation level $\sigma~=~0.1$.}
    \label{fig:generated_perturbed_sample}
\end{figure}

\clearpage

\begin{figure}[t]
    \centering
    \includegraphics[width=0.49\textwidth]{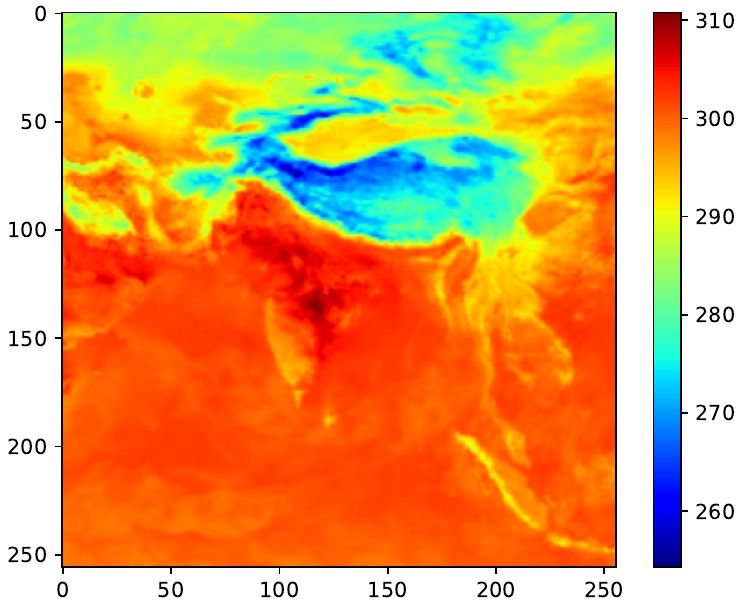}
    \hfill
    \includegraphics[width=0.49\textwidth]{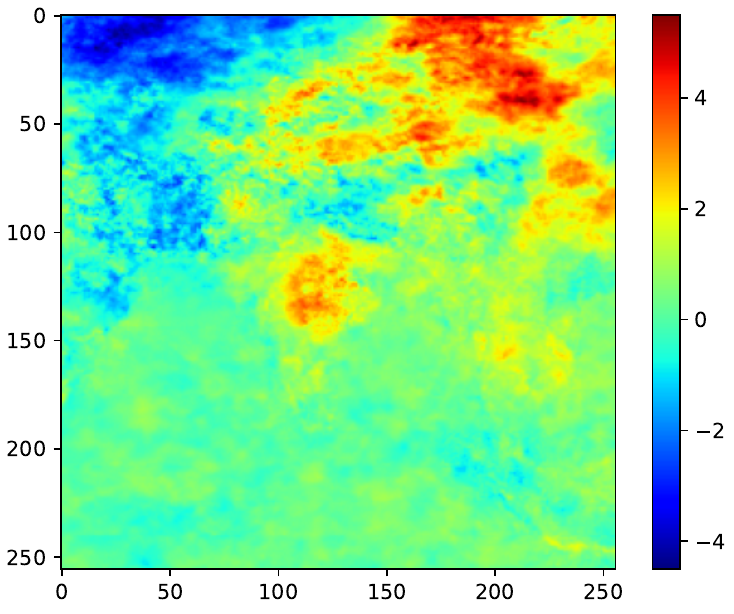}
    \includegraphics[width=0.49\textwidth]{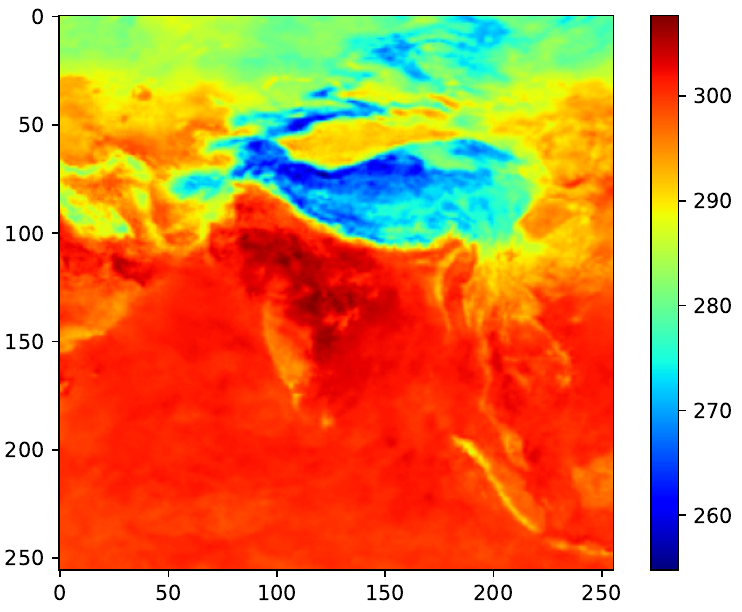}
    \hfill
    \includegraphics[width=0.49\textwidth]{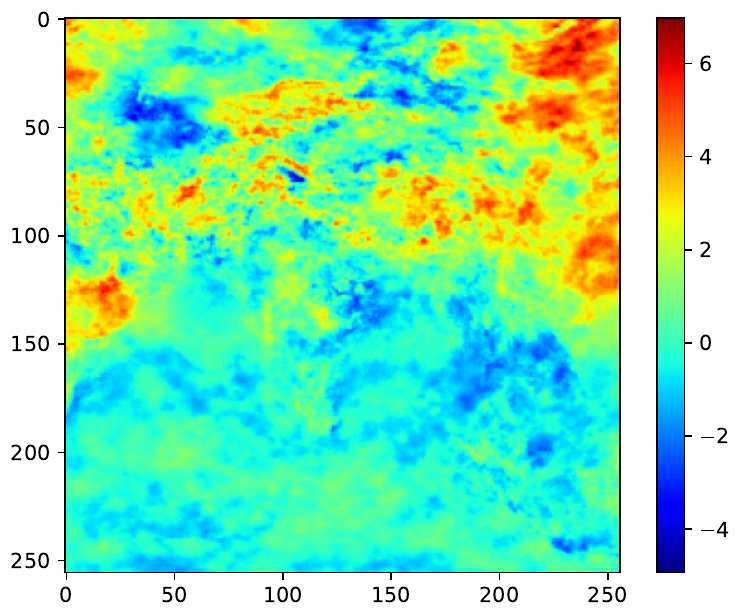}
    \includegraphics[width=0.9\textwidth]{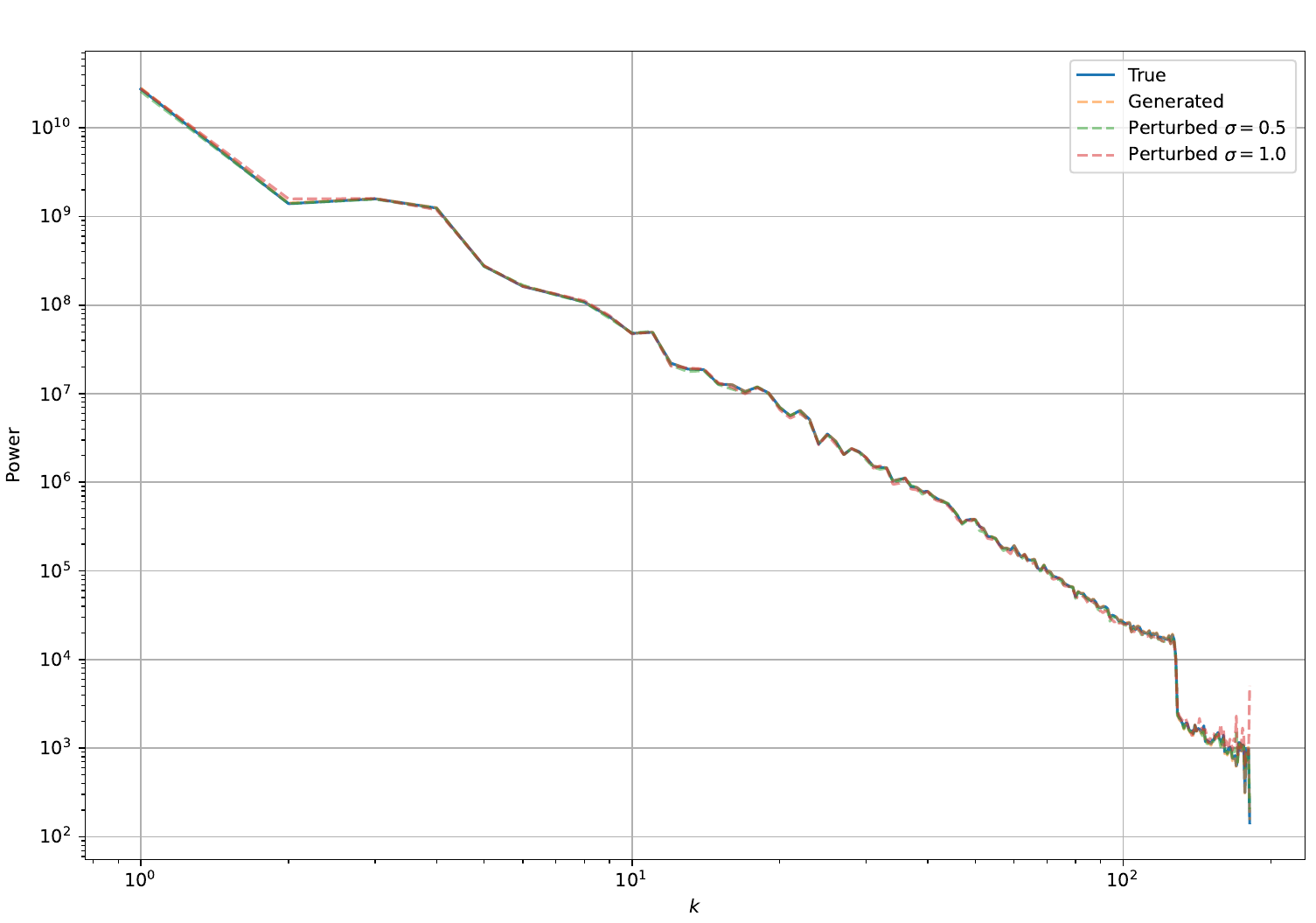}
    \caption{A perturbation of a true reanalysis plot of temperature field in $K$ (top left) and temperature perturbation field in $K$ for the 154th day-of-the-year at 00 UTC (top right) for perturbation level $\sigma = 0.5$. A perturbation of a true reanalysis plot of temperature field in $K$ (middle left) and temperature perturbation field in $K$ for the same day-of-the-year and same time of the day (middle right) for perturbation level $\sigma~=~1.0$. The power spectral density curves of temperature vs wavenumber($k$) for both these cases are compared with the real spectrum (bottom).}
    \label{fig:generated_perturbed_sample}
\end{figure}

\clearpage
\begin{figure*}[t]
    \centering
    \begin{tabular}{cccc}
        \subfloat[00 UTC Mean]{%
            \includegraphics[width=0.22\textwidth]{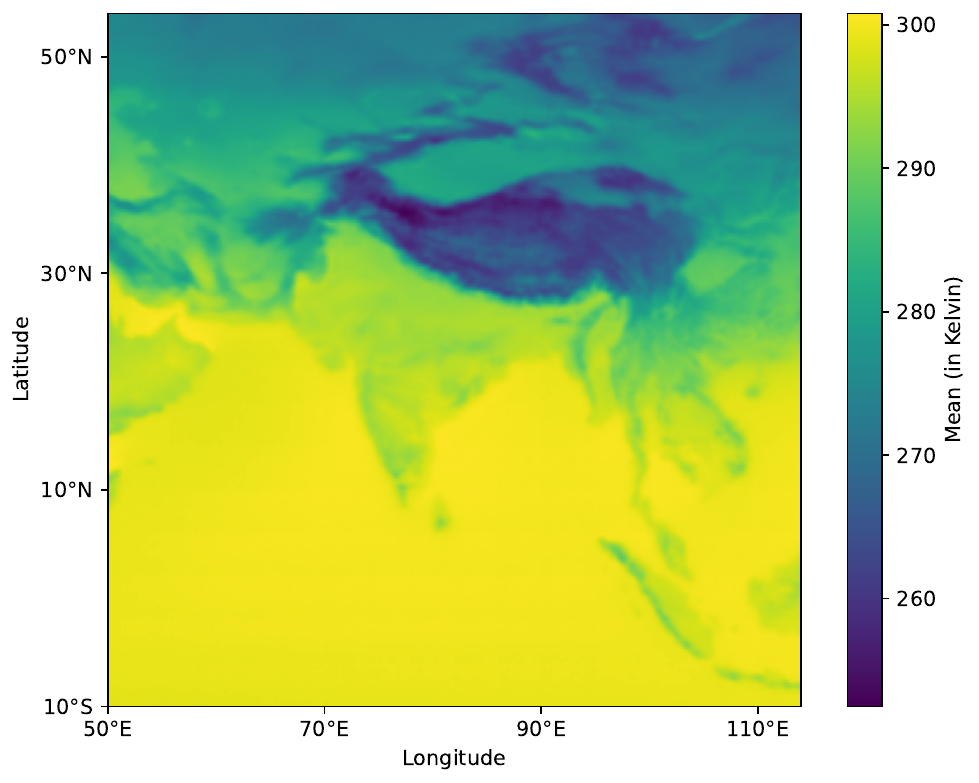}} &
        \subfloat[00 UTC Std.]{%
            \includegraphics[width=0.22\textwidth]{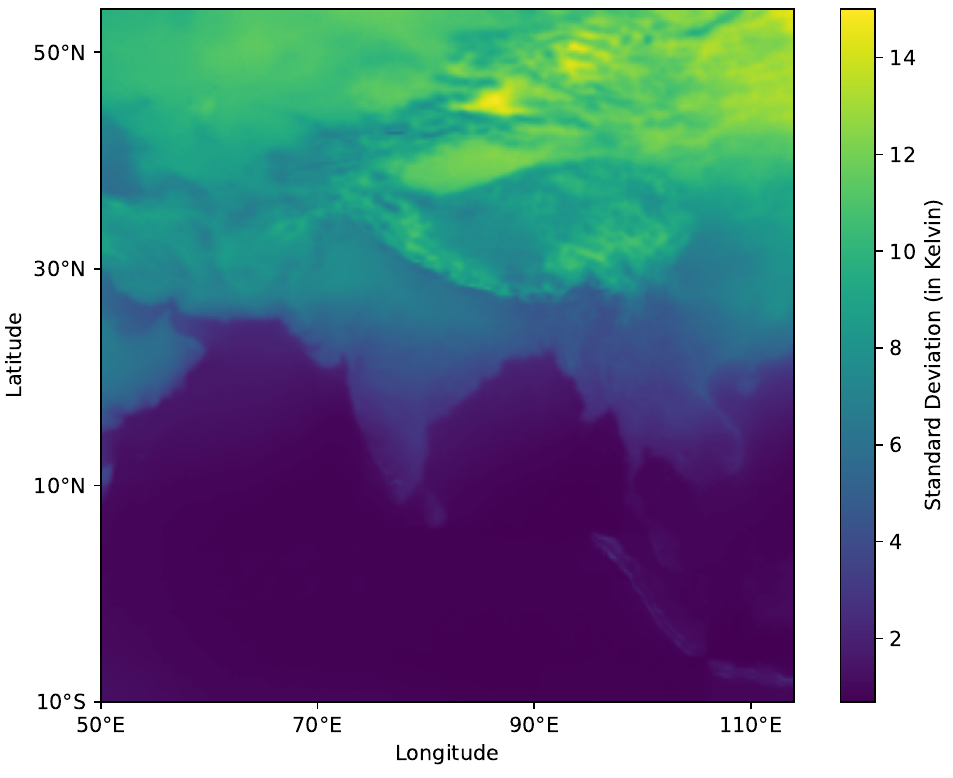}} &
        \subfloat[03 UTC Mean]{%
            \includegraphics[width=0.22\textwidth]{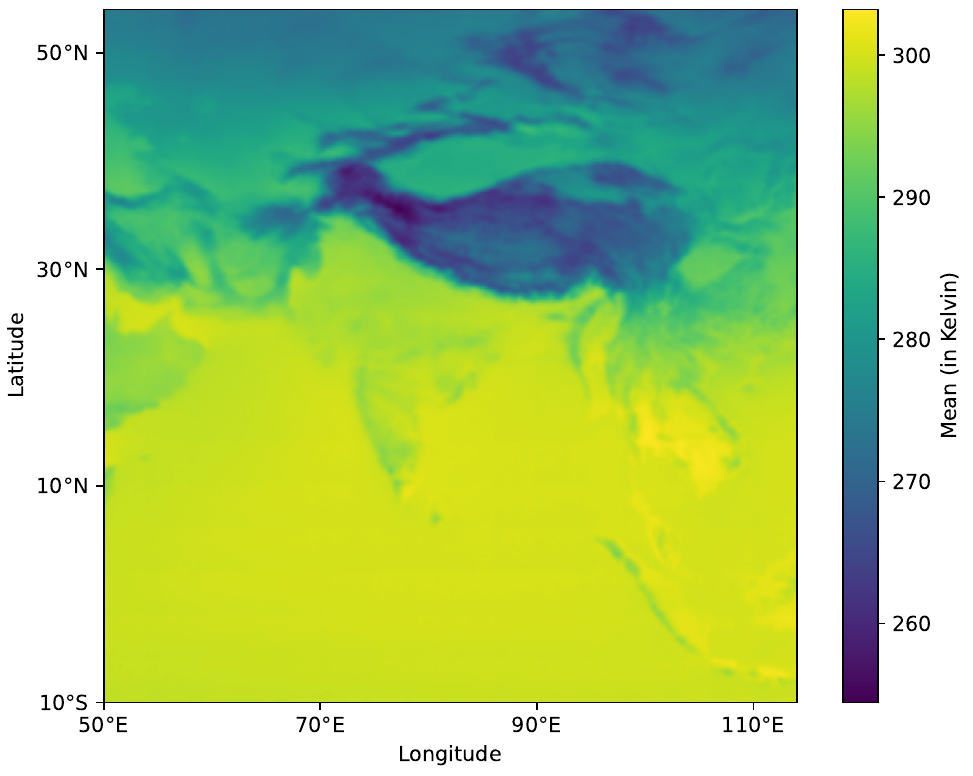}} &
        \subfloat[03 UTC Std.]{%
            \includegraphics[width=0.22\textwidth]{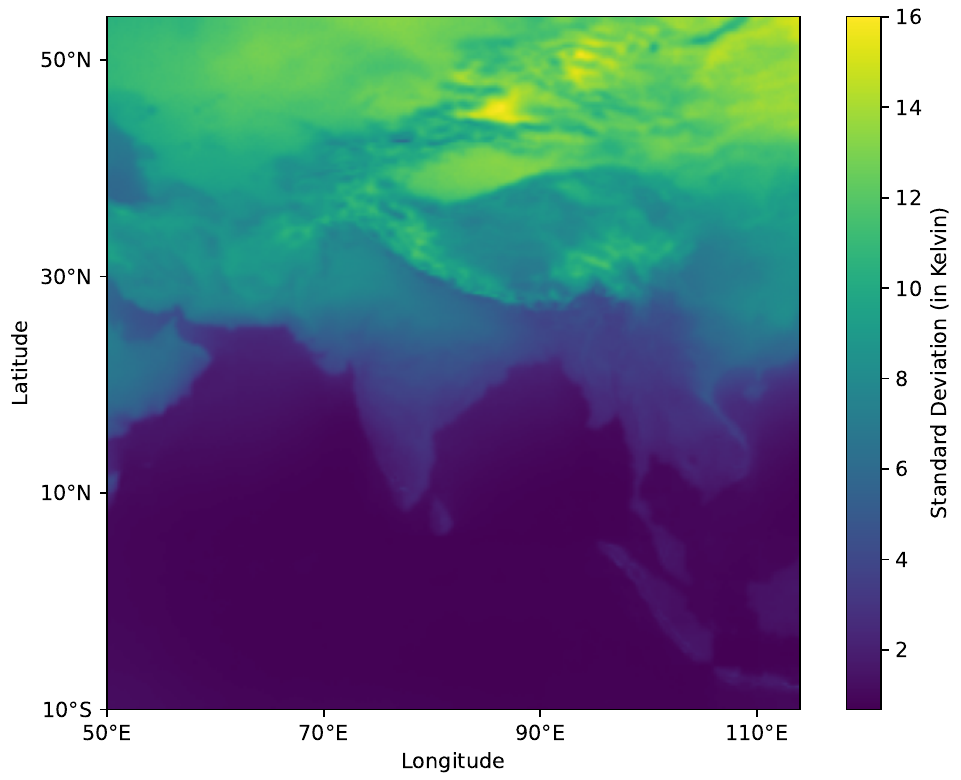}} \\[2pt]

        \subfloat[06 UTC Mean]{%
            \includegraphics[width=0.22\textwidth]{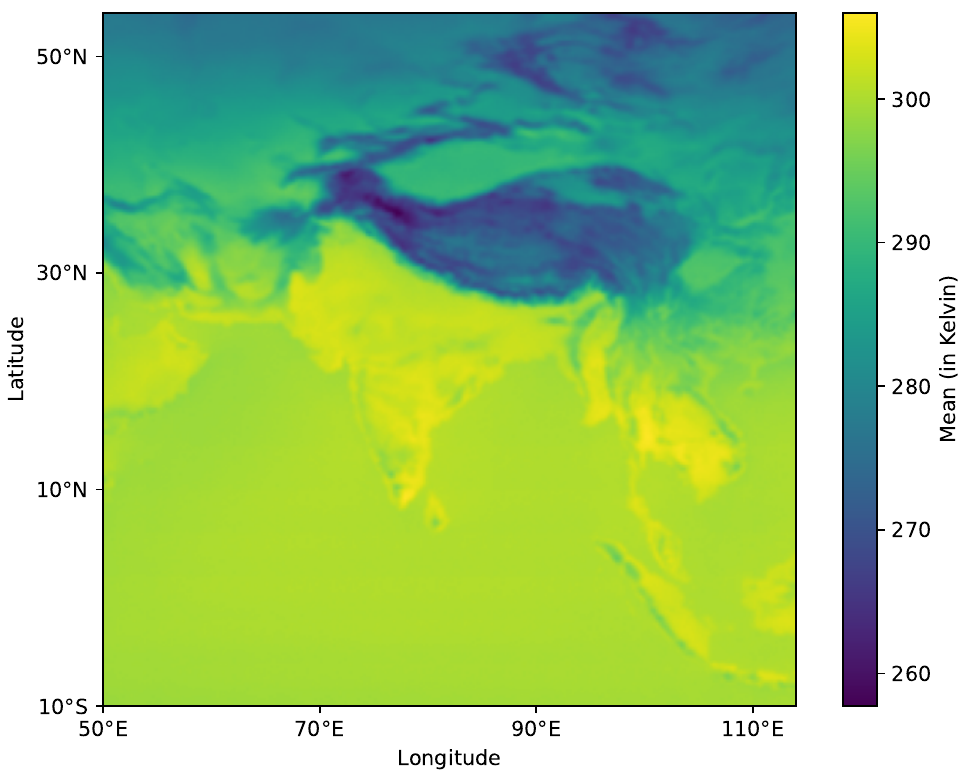}} &
        \subfloat[06 UTC Std.]{%
            \includegraphics[width=0.22\textwidth]{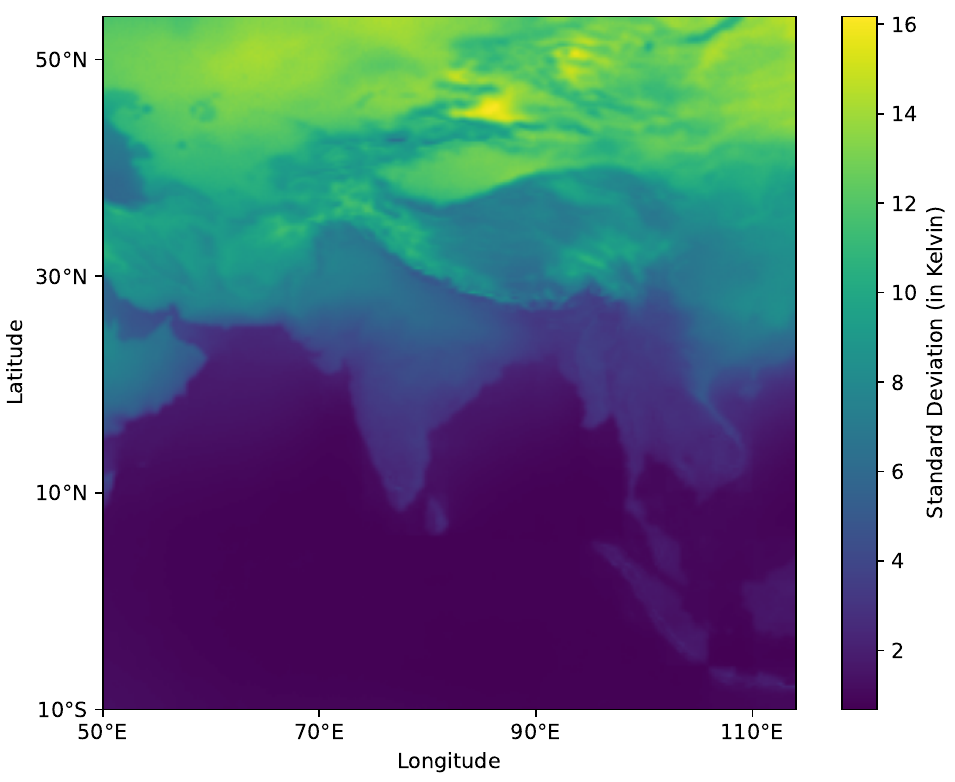}} &
        \subfloat[09 UTC Mean]{%
            \includegraphics[width=0.22\textwidth]{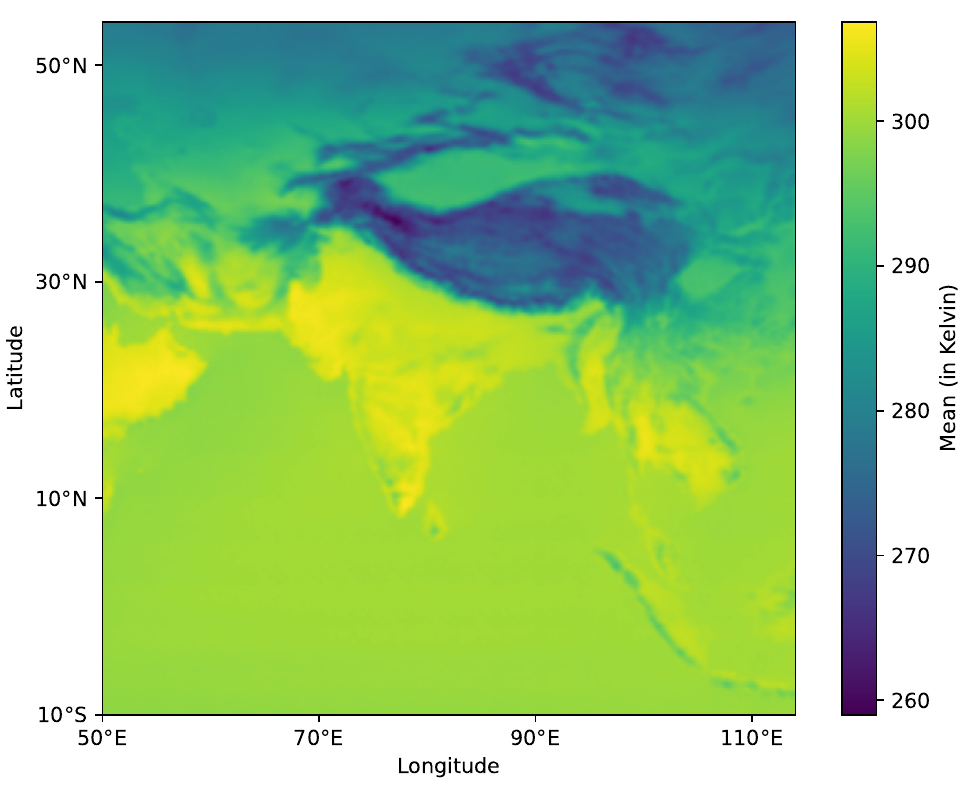}} &
        \subfloat[09 UTC Std.]{%
            \includegraphics[width=0.22\textwidth]{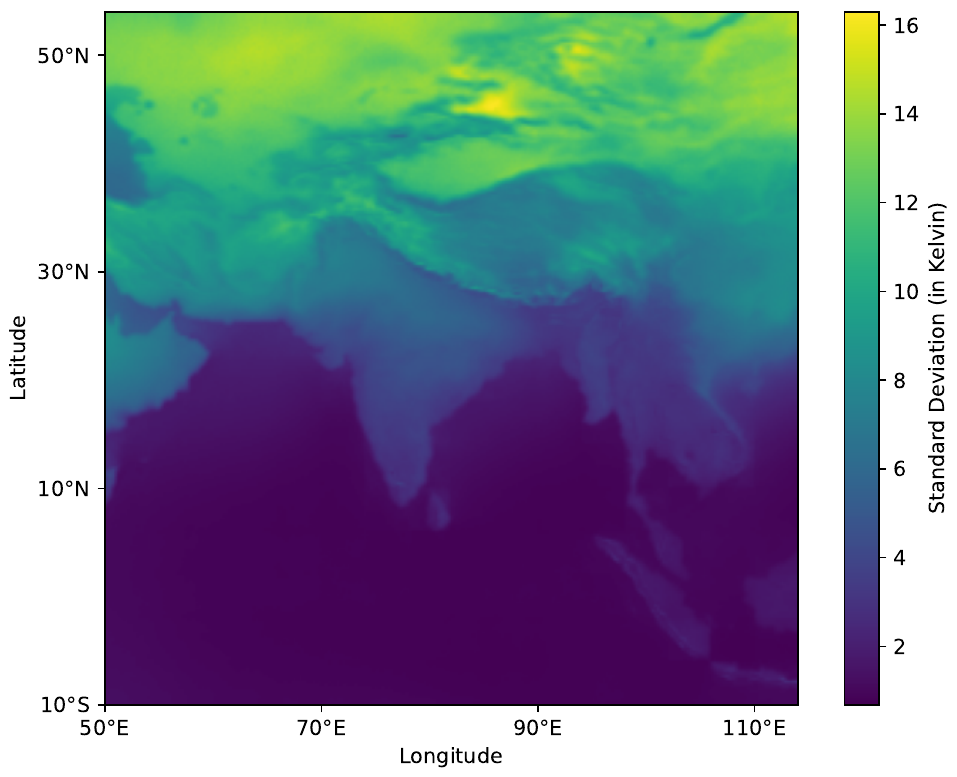}} \\[2pt]

        \subfloat[12 UTC Mean]{%
            \includegraphics[width=0.22\textwidth]{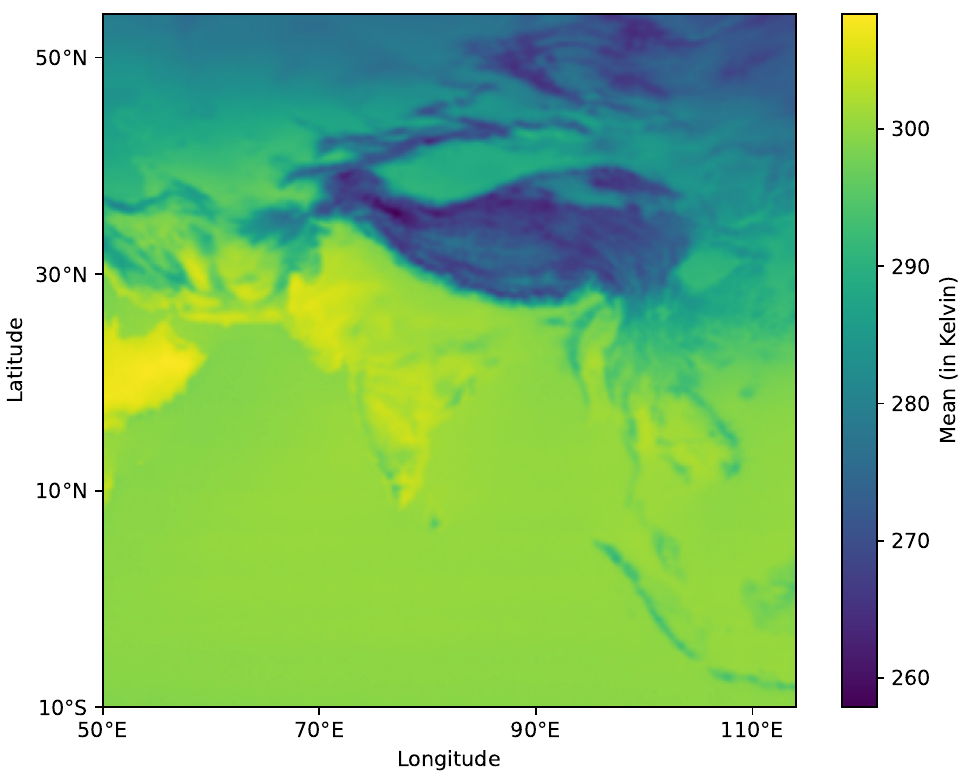}} &
        \subfloat[12 UTC Std.]{%
            \includegraphics[width=0.22\textwidth]{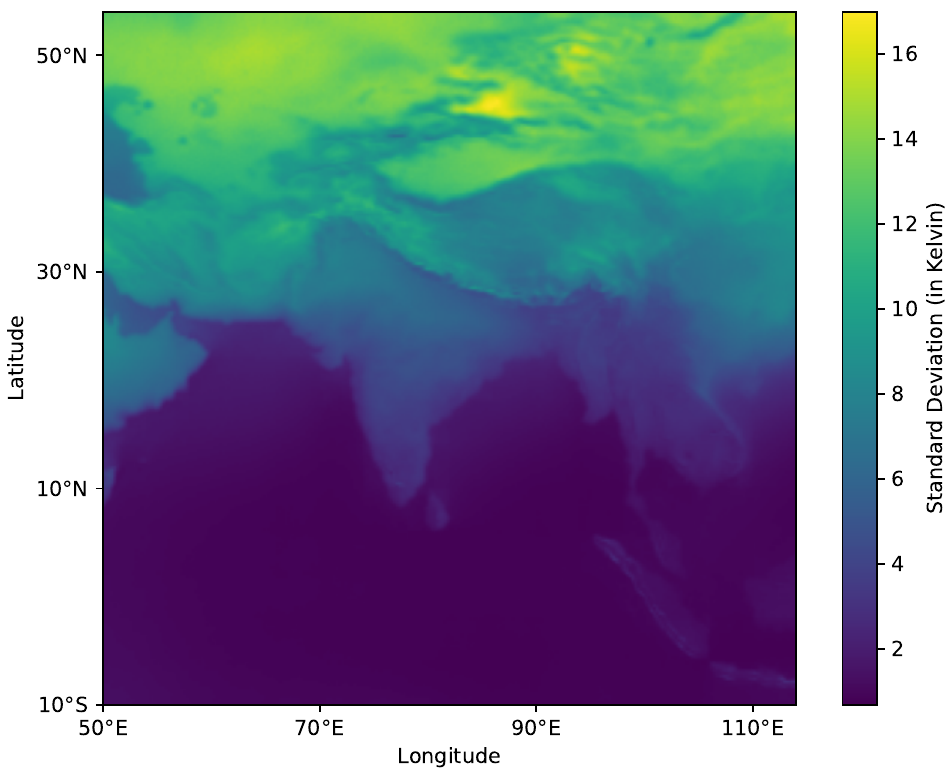}} &
        \subfloat[15 UTC Mean]{%
            \includegraphics[width=0.22\textwidth]{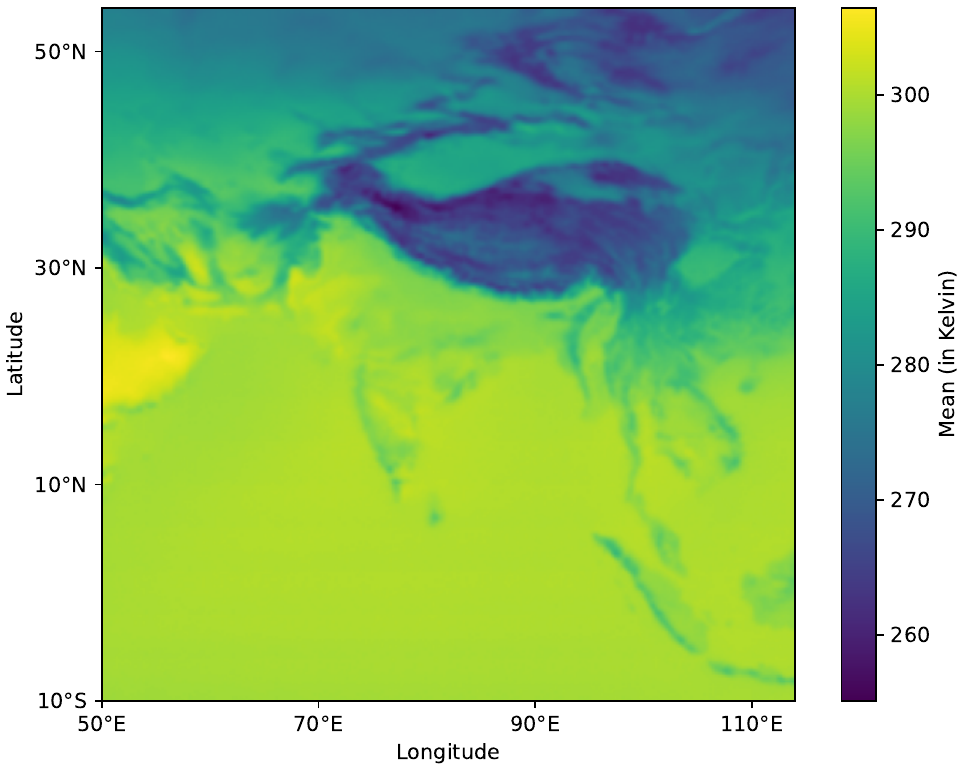}} &
        \subfloat[15 UTC Std.]{%
            \includegraphics[width=0.22\textwidth]{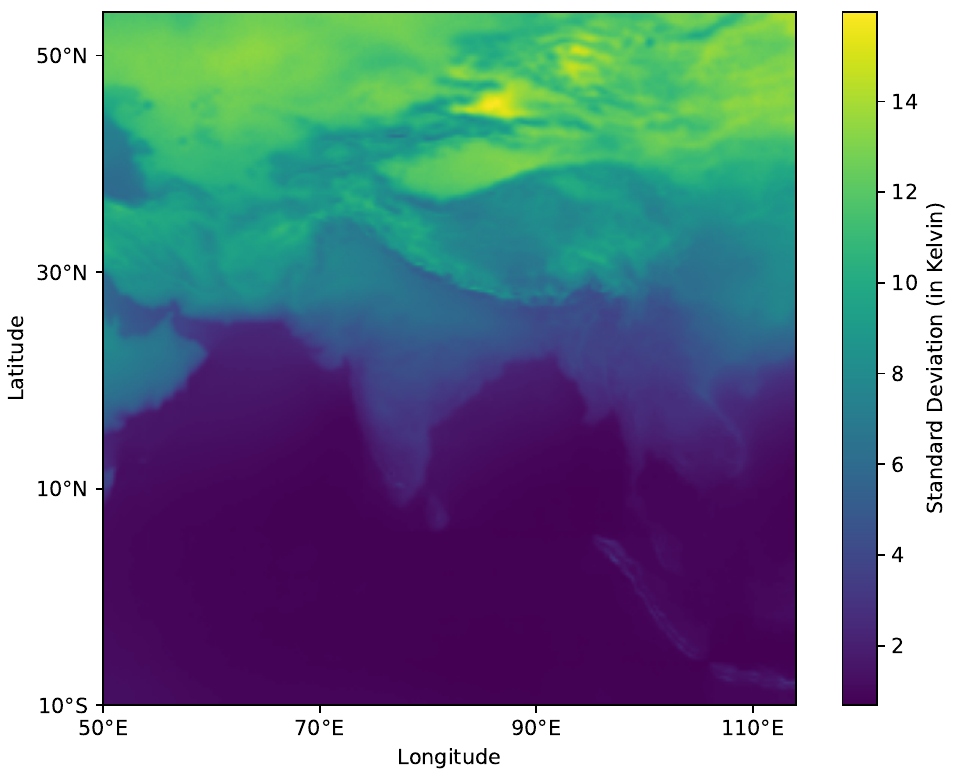}} \\[2pt]

        \subfloat[18 UTC Mean]{%
            \includegraphics[width=0.22\textwidth]{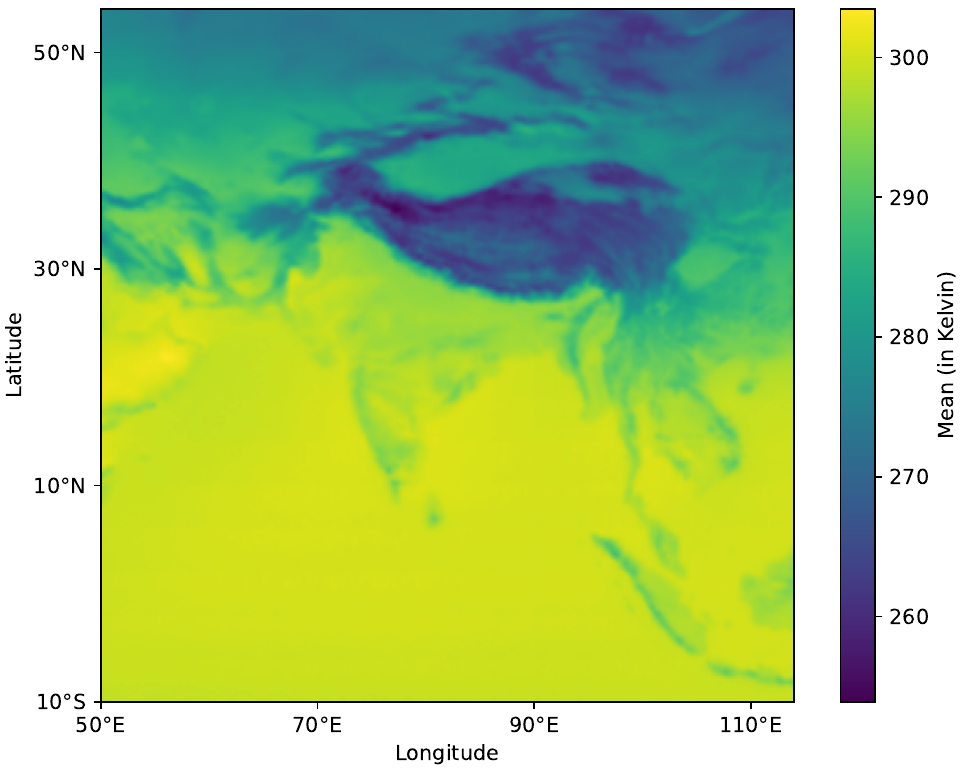}} &
        \subfloat[18 UTC Std.]{%
            \includegraphics[width=0.22\textwidth]{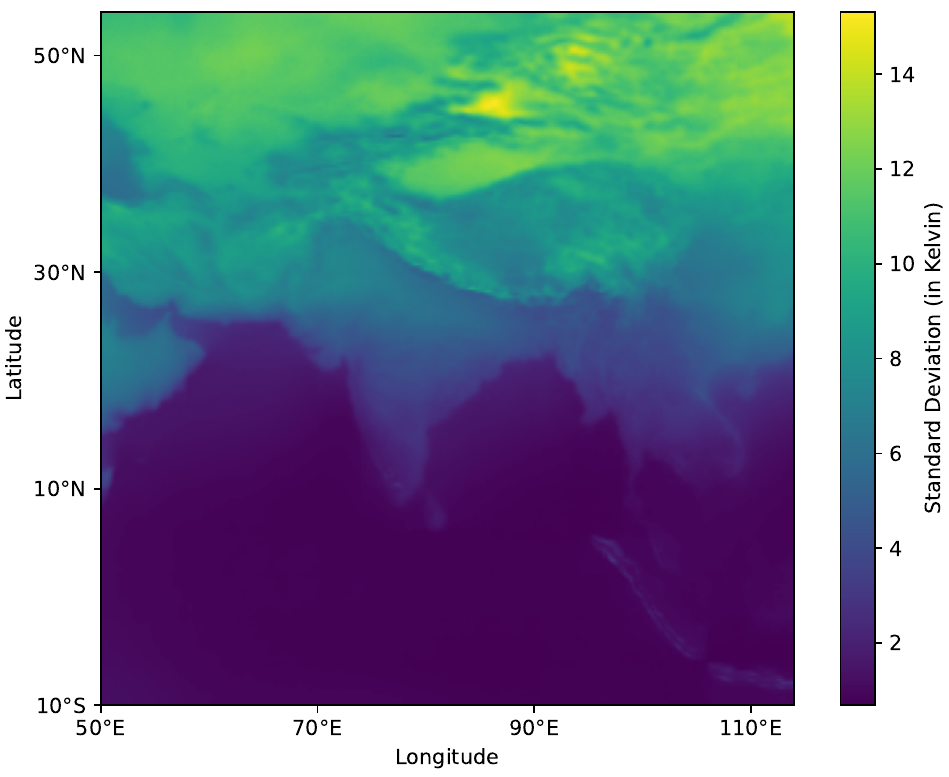}} &
        \subfloat[21 UTC Mean]{%
            \includegraphics[width=0.22\textwidth]{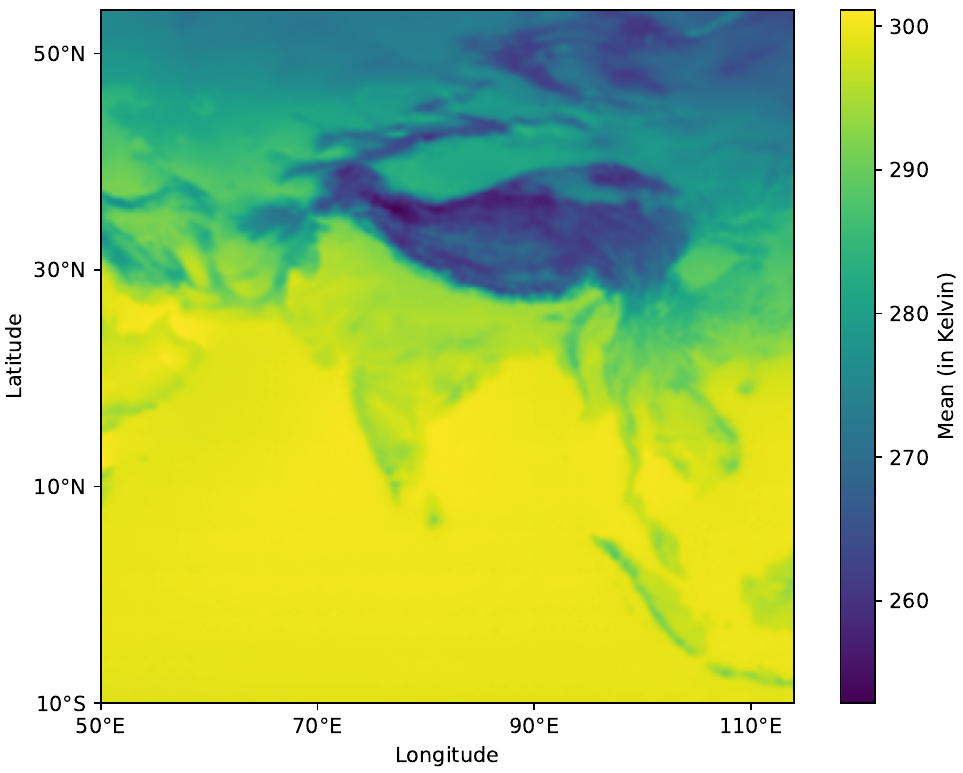}} &
        \subfloat[21 UTC Std.]{%
            \includegraphics[width=0.22\textwidth]{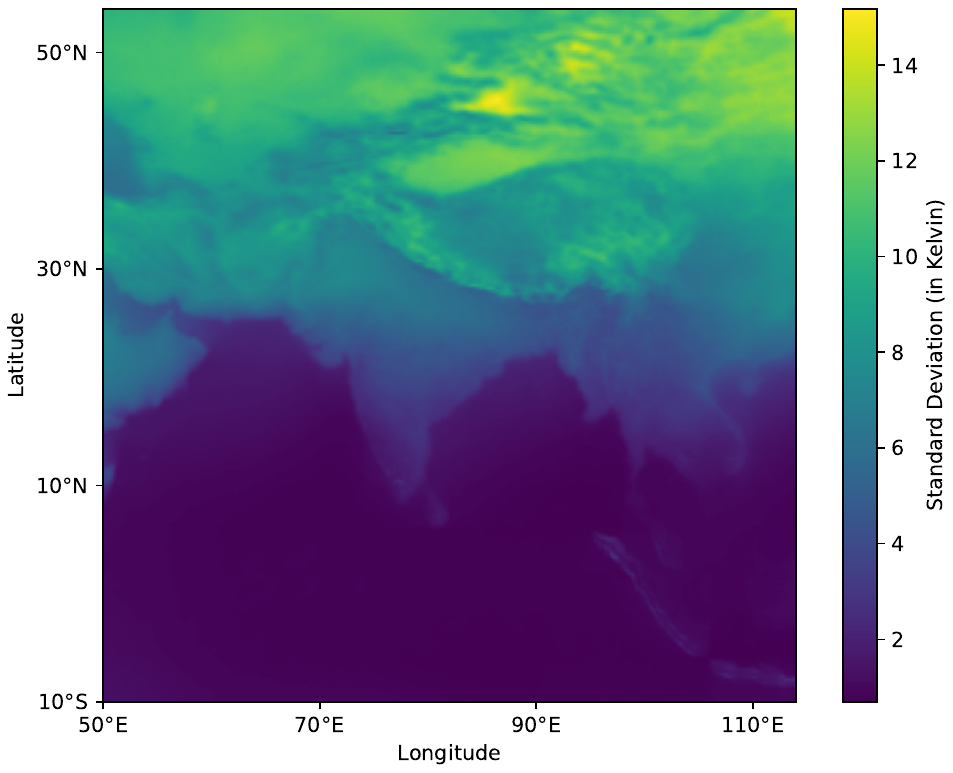}}
    \end{tabular}
    \caption{Historical climatological mean and standard deviation of 2-m air temperature at 3-hourly intervals.}
    \label{fig:historical_mean_std_i}
\end{figure*}

\subsection{Climate Variability}

In addition to the historical temperature fields, we provide precomputed climatological statistics for every day of the year and each 3-hourly time of day (00, 03, 06, 09, 12, 15, 18, and 21 UTC). These statistics are stored in a dictionary file, providing direct access to the seasonal and diurnal characteristics of temperature variability without requiring users to process the complete 40-year historical archive. For each day-of-year and 3-hourly time combination, the dictionary contains spatially resolved fields of the climatological mean (meantensor), standard deviation (stdtensor), expected maximum in 256 years (ExpectedMaxtensor), and expected minimum in 256 years (ExpectedMintensor). It also includes domain-wide summary statistics of the mean and standard deviation. Each spatial field retains the same $257\times257$ grid structure as the original temperature dataset, providing a compact representation of the spatial and temporal structure of temperature variability across the full geographic domain.

\begin{figure}
    \centering
    \includegraphics[width=0.95\linewidth]{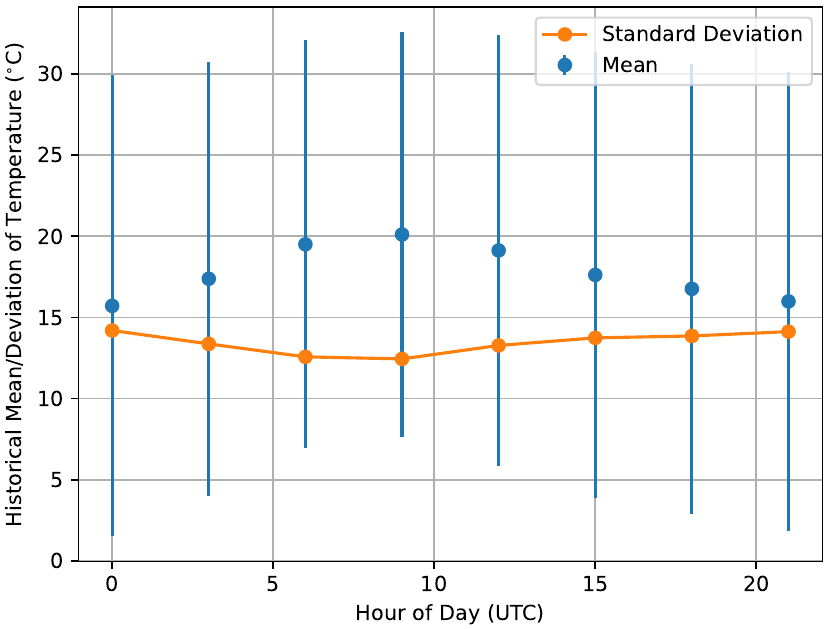}
    \caption{Hour-wise historical mean and standard deviation of the temperature field by sampling a large number of possible temperature fields using our generative model.}
    \label{fig:historical_mean_std_curve}
    
    \centering
    \includegraphics[width=0.48\linewidth]{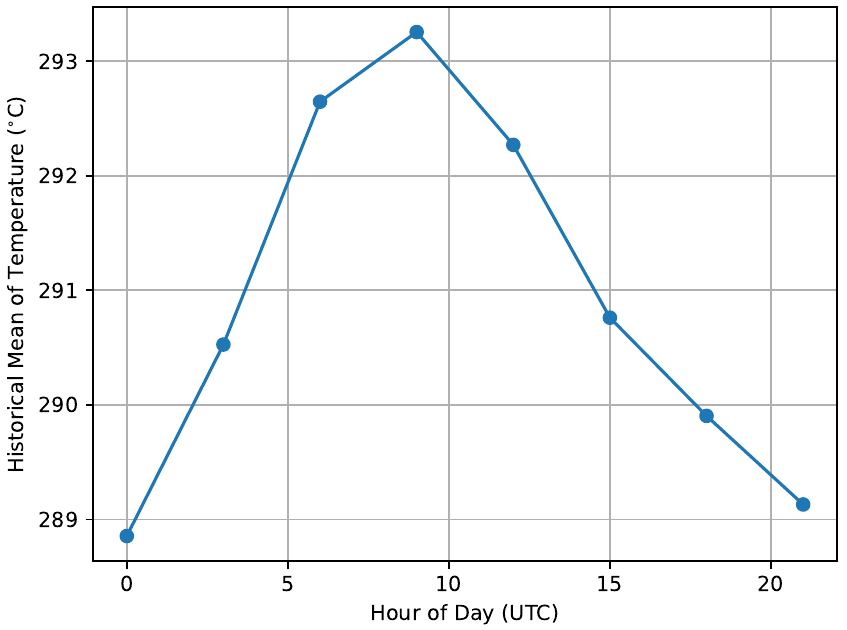} \hfill
    \includegraphics[width=0.48\linewidth]{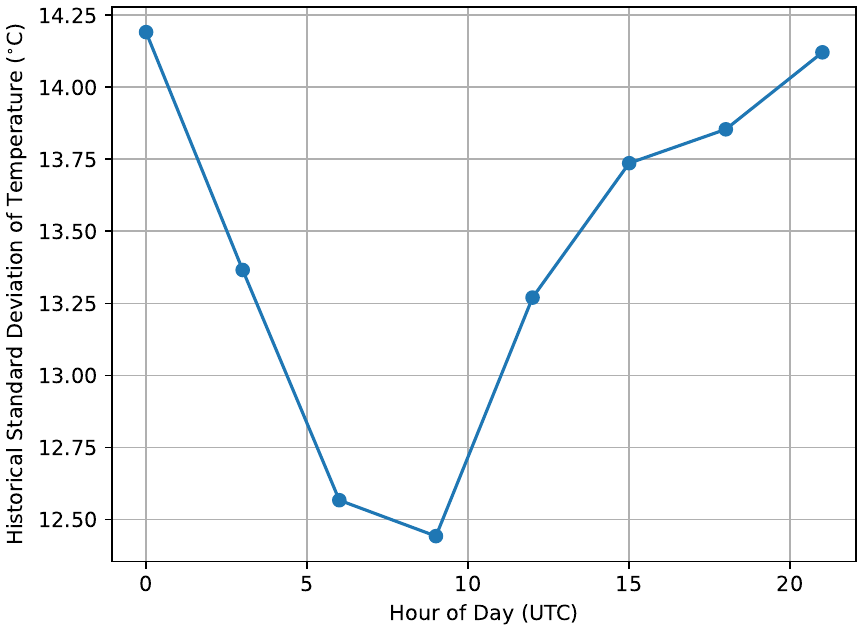}
    \caption{Hour-wise historical mean (left) and standard deviation (right) of the temperature field by sampling a large number of possible temperature fields using our generative model.}
    \label{fig:historical_mean_std_curve2}

    \centering
    \includegraphics[width=\linewidth]{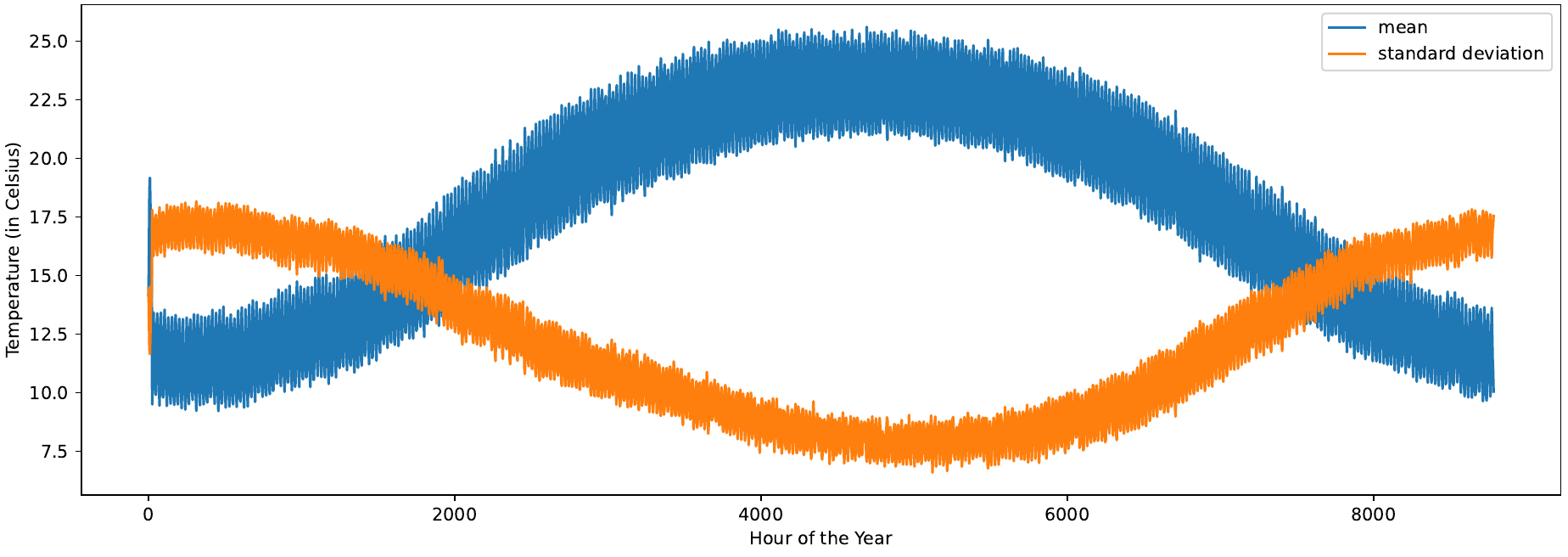} 
    \caption{Overall historical variability over the region estimated every three hours over the year.}
    \label{fig:variability}
\end{figure}

Temperature variability across the region exhibits substantial seasonal and diurnal dependence, arising from differences in climate, topography, land-surface characteristics, and atmospheric circulation. Resolving temperature statistics according to both day of year and time of day therefore provides a direct representation of how the magnitude and spatial structure of temperature fluctuations evolve throughout the annual and diurnal cycles. For each temporal condition, the historical dataset is used to calculate spatially varying climatological mean and standard deviation fields. The climatological mean represents the typical temperature state at each spatial location, while the standard deviation characterizes the magnitude of temperature fluctuations around that state. Examples of these climatological mean and standard deviation fields for different times of day are shown in \Cref{fig:historical_mean_std_curve,fig:historical_mean_std_curve2}.

We further employ conditional flow matching to generate temperature variability fields conditioned on a specified day of year and 3-hourly time of day. By conditioning the generative model on these temporal coordinates, the generated fields capture the seasonal and diurnal structure learned from the historical record in ERA5. The model can produce thousands of realizations for a given temporal condition, enabling characterization of the spatial distribution and range of temperature variability beyond the finite set of samples available in the historical archive.

For each spatial location, the climatological standard deviation provides an estimate of the typical magnitude of temperature fluctuations associated with a particular day of year and 3-hourly time of day. This temporal organization provides a useful framework for conditioning, evaluating, and interpreting temperature variability and is more informative than assuming a temporally uniform variance. It also allows regional differences in both the magnitude and temporal structure of temperature variability to be examined systematically.

The resulting variability fields provide an additional representation of temperature variability beyond the individual historical temperature fields. The 40-year dataset provides the empirical basis for learning spatial temperature structure, while the day-of-year and 3-hourly climatology provides the temporal framework for characterizing variability. Together, these components establish a reusable framework for climate analysis, machine-learning development, ensemble generation, uncertainty characterization, and climate-data augmentation. The overall variability averaged over the whole region can be seen in \Cref{fig:variability}. Similar charts can be obtained for each grid or a couple of grids to represent a city's temperature variability, using the already estimated temperature variability fields. 

Overall, the two generative applications provide complementary information. The first generates spatially coherent temperature perturbations that can be applied to reference temperature fields to create additional realizations. The second uses temporal conditioning to characterize temperature variability for a specified day of year and 3-hourly time of day. Together with the precomputed climatological statistics stored in the dictionary file, these components extend the utility of the historical dataset beyond direct analysis of the original reanalysis fields and provide a framework for investigating the spatial and temporal structure of temperature variability.

\section{Discussion}

The 40-year temperature history provides a consistent regional representation of 2-m air temperature across a geographically and climatically diverse domain. Its long temporal coverage, 3-hourly sampling, and $0.25^\circ$ spatial resolution support analysis of seasonal, diurnal, and spatial temperature variability. By providing a predefined regional subset in a machine-learning-oriented format, the dataset also reduces the need for repeated extraction and preprocessing of large global reanalysis archives and facilitates reproducible AI/ML experimentation.

An important component of the dataset is the set of precomputed climatological statistics conditioned on day of year and 3-hourly time of day. These statistics provide a compact representation of the seasonal and diurnal temperature cycle and can be used directly for climatological analysis, normalization, conditional model development, and evaluation of generated temperature fields. Resolving the statistics in both space and time also preserves regional differences in the magnitude and structure of temperature variability that would be lost under a temporally uniform climatological representation.

The generative-modeling experiments provide a demonstration of the dataset's utility for learning spatially structured temperature variability. The flow-matching model generates temperature fields and perturbations that exhibit spatial organization learned from the historical data. The similarity between reference and generated power spectral density provides evidence that the model captures aspects of the scale-dependent spatial structure of temperature variability. Conditional generation further demonstrates that the temporal information provided by the dataset can be incorporated into a generative framework to produce realizations corresponding to specified day-of-year and time-of-day conditions. These results illustrate one possible use of the dataset rather than constituting the primary contribution of the study.

Several limitations should be considered. First, the temperature fields inherit the uncertainties, assumptions, and spatial representation of the ERA5 reanalysis. The dataset should therefore be regarded as a processed regional reanalysis resource rather than an observational dataset. Second, the $0.25^\circ$ spatial resolution limits the representation of fine-scale processes and local temperature variability, particularly in complex terrain and densely urbanized areas. Third, the generative models are trained on the statistical characteristics of the historical dataset and therefore cannot be expected to reproduce variability outside the learned distribution without additional conditioning or training data. Generated fields should consequently be interpreted as statistical realizations of temperature variability rather than forecasts of specific future atmospheric states.

The chronological training and testing split provides an out-of-sample evaluation period, but additional evaluation against independent observations or reanalysis products would strengthen assessment of generalization. In particular, comparisons of spatial correlations, temporal distributions, spectral characteristics, extremes, and other physically relevant statistics could provide a more comprehensive evaluation of generated temperature fields. Such evaluation is also important before using generated samples for applications requiring calibrated uncertainty estimates or quantitative probabilistic interpretation.

\subsection{Future Extensions}

The dataset provides a foundation for several extensions of regional atmospheric machine-learning research. Future versions could incorporate additional atmospheric and surface variables, enabling multivariate generation and investigation of relationships between temperature and other components of the atmospheric state. Higher-resolution datasets could also be developed for applications requiring finer representation of topographic, coastal, or urban temperature variability.

The conditioning framework could be extended to include geographic information, large-scale atmospheric states, circulation regimes, and other environmental predictors. Such extensions could allow generative models to represent more event-specific temperature variability and could provide a pathway toward applications such as heatwave analysis, statistical downscaling, and climate-model evaluation.

Future studies should also evaluate generated samples against independent datasets using a broader set of physical and statistical diagnostics, including spatial and temporal correlations, temperature distributions, spectral characteristics, persistence, and extreme-temperature statistics. These evaluations would help determine the extent to which generative models trained on the dataset reproduce physically relevant variability and would provide a stronger basis for applications involving ensemble generation and uncertainty quantification.

\section{Conclusions}

We used a 40-year, 3-hourly, $0.25^\circ$ gridded sub-sampled dataset of 2-m air temperature covering South Asia and surrounding regions for 1985--2024 from the ERA5 dataset. The data subset provides a consistently processed regional temperature archive together with machine-learning-oriented data organization, a predefined chronological training and testing split, and precomputed climatological statistics resolved by day of year and 3-hourly time of day. These characteristics make the data subset suitable for reproducible climate analysis and AI/ML experimentation without requiring users to repeatedly process the corresponding large-scale global reanalysis archive.

We demonstrate the utility of the dataset through two complementary generative-modeling applications based on flow matching. The first generates spatially coherent temperature perturbations, while the second generates temperature realizations conditioned on seasonal and diurnal information. The generated fields reproduce aspects of the spatial organization and scale-dependent variability represented in the historical temperature data, illustrating that the dataset contains sufficient structure for data-driven generative modeling.

The primary contribution of this work is the regional, ML-ready temperature dataset and its associated climatological resources. The generative experiments provide an example of how these resources can be used to develop and evaluate models for spatially structured atmospheric variability. The dataset can therefore serve as a reusable foundation for regional climate analysis, spatiotemporal machine learning, generative modeling, data augmentation, and related AI-based research. Future extensions to additional variables, higher spatial resolutions, richer conditioning information, and independent evaluation datasets could further broaden its utility for weather and climate applications.

\section*{Availability Statement}

The dataset and codes for preprocessing, analysis, and reproduction of the reported results used in this study, \emph{Indian Subcontinent 2m Temperature Field}, is publicly available on Kaggle: \url{https://www.kaggle.com/datasets/siddharthrout/indian-subcontinent-temperature-field}. The exact dataset version used in this study is \textbf{2026-10-05} and license \textbf{Creative Commons Attribution 4.0 International (CC BY 4.0)}.

\section*{Conflict of Interest}
The authors declare there are no conflicts of interest for this manuscript.

\acknowledgments
We acknowledge the suggestions and insightful discussions with Shoyon Panday and St\'ephane Gaudreault.
% We acknowledge and we are grateful for the funding from Environment and Climate Change Canada through grant GCXE25S053.

%%%%%%%%%%%%%%%%%%%%%%%%%%%%%%%%%%%%%%%%%%%%%%%
% REFERENCES and BIBLIOGRAPHY
%
% \bibliography{<name of your .bib file>} don't specify the file extension
% don't specify bibliographystyle
%
%%%%%%%%%%%%%%%%%%%%%%%%%%%%%%%%%%%%%%%%%%%%%%%
% \bibliographystyle{unsrt}
\bibliography{refs}

%Reference citation instructions and examples:
%
% Please use ONLY \cite and \citeA for reference citations.
% \cite for parenthetical references
% ...as shown in recent studies (Simpson et al., 2019)
% \citeA for in-text citations
% ...Simpson et al. (2019) have shown...
%
%
%...as shown by \citeA{jskilby}.
%...as shown by \citeA{lewin76}, \citeA{carson86}, \citeA{bartoldy02}, and \citeA{rinaldi03}.
%...has been shown \cite{jskilbye}.
%...has been shown \cite{lewin76,carson86,bartoldy02,rinaldi03}.
%... \cite <i.e.>[]{lewin76,carson86,bartoldy02,rinaldi03}.
%...has been shown by \cite <e.g.,>[and others]{lewin76}.
%
% apacite uses < > for prenotes and [ ] for postnotes
% DO NOT use other cite commands (e.g., \citet, \citep, \citeyear, \nocite, \citealp, etc.).
%

\end{document}